\documentclass[journal,twoside]{IEEEtran}

\usepackage{graphicx}
\usepackage{subcaption}  
\usepackage{romannum}
\usepackage{empheq}
\usepackage{amsmath}
\usepackage{amsfonts}
\usepackage{amssymb}
\usepackage{amsthm}
\usepackage{lipsum}
\usepackage{caption}
\usepackage{bm}
\usepackage{mathtools}
\usepackage[noadjust]{cite}
\usepackage{float}
\usepackage{tikz}
\usetikzlibrary{arrows.meta, positioning, fit, calc}

\usepackage[ruled,vlined]{algorithm2e}
\usepackage[noend]{algpseudocode}
\usepackage{romannum}

\usepackage{color}

\title{Lightweight Edge Learning via Dataset Pruning}

 \author {~Laha Ale,~Hu Luo,~Mingsheng Cao,~Shichao Li,~Huanlai Xing and~Haifeng Sun

\thanks{L. Ale, H. Luo and H. Xing are with the School of Computing and Artificial Intelligence, Southwest Jiaotong University, Chengdu, China (e-mail:\{luohu2025@my., hxx@home. and laha\_ale@\} swjtu.edu.cn).}%

\thanks{M. Cao is with the Network and Data Security Key Laboratory of Sichuan Province, University of Electronic Science and Technology of China, Chengdu, China (e-mail: cms@uestc.edu.cn).}%

\thanks{S. Li is with the School of Information and Communication, Guilin University of Electronic Technology, Guilin, China (e-mail: shichaoli@guet.edu.cn).}
\thanks{H. Sun is with the School of Computer Science and Technology, the Southwest University of Science and Technology, Mainyang, China (e-mail: sunhaifeng@swust.edu.cn).}

 }

\begin{document}

\maketitle

\begin{abstract} 
Edge learning facilitates ubiquitous intelligence by enabling model training and adaptation directly on data-generating devices, thereby mitigating privacy risks and communication latency. However, the high computational and energy overhead of on-device training hinders its deployment on battery-powered mobile systems with strict thermal and memory budgets. While prior research has extensively optimized model architectures for efficient inference, the training phase remains bottlenecked by the processing of massive, often redundant, local datasets. In this work, we propose a data-centric optimization framework that leverages dataset pruning to achieve resource-efficient edge learning. Unlike standard methods that process all available data, our approach constructs compact, highly informative training subsets via a lightweight, on-device importance evaluation. Specifically, we utilize average loss statistics derived from a truncated warm-up phase to rank sample importance, deterministically retaining only the most critical data points under a dynamic pruning ratio. This mechanism is model-agnostic and operates locally without inter-device communication. Extensive experiments on standard image classification benchmarks demonstrate that our framework achieves a near-linear reduction in training latency and energy consumption proportional to the pruning ratio, with negligible degradation in model accuracy. These results validate dataset pruning as a vital, complementary paradigm for enhancing the sustainability and scalability of learning on resource-constrained mobile edge devices.
\end{abstract}


\IEEEpeerreviewmaketitle
\thispagestyle{plain} 
\pagestyle{plain}    
\pagenumbering{arabic}   
\section{Introduction}
\label{sec_intro}

The proliferation of intelligent sensing and decision-making applications has driven a paradigm shift from centralized cloud computing toward edge learning~\cite{Ale2024}. By performing data processing directly at the source, edge learning supports real-time applications in Internet-of-Things (IoT) systems, autonomous platforms, and cyber--physical systems~\cite{9955525}. However, unlike cloud environments with abundant resources, edge devices operate under stringent constraints regarding computation capability, energy budget, memory, and storage~\cite{9757749,sun2024EC}. These constraints fundamentally alter the design requirements for learning algorithms deployed in the wild.

Existing efforts on efficient edge learning have predominantly focused on reducing the \emph{model-side} cost. Techniques such as model compression, network pruning~\cite{Iterative_Pruning2017,10633894_pruning,shim2024snp}, quantization~\cite{zhou2017incremental,jacob2018quantization,qu2025automatic}, and lightweight architecture design~\cite{szegedy2015Googlenet,howard2017mobilenets,zhang2018Shufflenet,sandler2018mobilenetv2} aim to reduce the complexity of the neural network itself. While these approaches effectively reduce inference latency, the overhead of \emph{training and on-device adaptation} remains a critical bottleneck. In many edge scenarios, models must be trained or fine-tuned locally to adapt to dynamic environments. Repeatedly processing large, raw datasets during this phase can quickly exceed device resource budgets. Consequently, the feasibility of edge learning is often limited not merely by the size of the model, but by the volume of data that must be processed to train it.

This observation motivates a complementary, \emph{data-centric} perspective on edge learning. Real-world data collected at the edge often exhibits substantial redundancy; many samples are highly correlated or easily learned~\cite{Toneva2019Forgetting}, providing diminishing returns for model performance improvement. An essential question therefore arises: \emph{which data samples are worth processing under strict resource constraints?} Addressing this requires rethinking learning pipelines to explicitly control data usage, rather than relying on the conventional assumption that all available local data must be utilized.

In this work, we investigate dataset pruning as a core mechanism for enabling resource-efficient edge learning. Unlike offline pruning in centralized settings, we formulate dataset pruning as a local, on-device decision process. By selecting compact yet informative subsets of local data prior to full training, edge devices can substantially reduce training iterations, latency, and energy consumption. Importantly, this data-centric approach is orthogonal to model-centric efficiency techniques and can be seamlessly combined with them to further optimize performance.

Implementing dataset pruning on edge devices, however, presents unique challenges. The joint optimization of data selection and model parameters is a combinatorial problem that is computationally intractable. Furthermore, the evaluation of sample importance must be lightweight and conducted entirely on-device; edge nodes cannot afford expensive second-order analysis or full training convergence strictly for scoring purposes. To address these challenges, we propose a framework that leverages lightweight statistics obtained from a short warm-up training phase. Each device independently assigns importance scores to its local samples and deterministically retains only the most informative data under a prescribed pruning ratio. The resulting pruned dataset is then utilized for standard training without modifying the underlying optimization algorithm.

The main contributions of this paper are summarized as follows:
\begin{itemize}
    \item We formulate dataset pruning as a resource-aware optimization problem for local edge learning, explicitly capturing the trade-off between learning performance and computational, energy, and storage costs.
    \item We propose an importance-based dataset pruning method that utilizes lightweight statistics from a truncated warm-up phase to identify high-value training samples with negligible overhead.
    \item We develop a deterministic pruned dataset construction mechanism that directly enforces resource constraints and integrates seamlessly with standard on-device learning pipelines.
    \item We provide a computational complexity analysis and empirical evaluation on standard benchmarks, demonstrating that the proposed approach achieves near-linear reductions in training cost with minimal degradation in model accuracy.
\end{itemize}

The remainder of this paper is organized as follows. Section~\ref{sec_system} introduces the system model and characterizes the associated resource costs. Section~\ref{sec_problem} formulates the constraint-aware dataset pruning problem. Section~\ref{sec_method} details the proposed importance-based pruning algorithm. Section~\ref{sec_results} presents the experimental evaluation and performance analysis. Finally, Section~\ref{sec_conclusion} concludes the paper.

\section{Related Work}
\label{sec_related}

This work lies at the intersection of efficient edge learning and data-centric optimization. We review the related literature from three perspectives: (i) efficient learning methods for edge computing, (ii) dataset pruning and coreset selection, and (iii) data-centric learning approaches.

\subsection{Efficient Learning for Edge Computing}

Edge computing~\cite{2020ECsurvey,9757749,sun2024EC,Ale2022EC} has emerged as a promising paradigm for enabling low-latency, privacy-preserving intelligence by pushing learning and inference closer to data sources. A large body of research has focused on reducing the computational and energy cost of learning on resource-constrained edge devices. Representative approaches include lightweight model architectures~\cite{szegedy2015Googlenet,howard2017mobilenets,zhang2018Shufflenet,sandler2018mobilenetv2}, network pruning~\cite{Iterative_Pruning2017,10633894_pruning,shim2024snp}, quantization~\cite{zhou2017incremental,jacob2018quantization,qu2025automatic}, low-rank factorization~\cite{2020LowRank}, and knowledge distillation~\cite{park2019relational,Beyer2022Kd}. These model-centric techniques aim to reduce the complexity of neural networks while maintaining acceptable accuracy.

In parallel, federated learning~\cite{FederatedL&EC,XIAO2021107338,lim2020FederatedLinEC} and distributed optimization frameworks enable collaborative model training across edge devices without sharing raw data. Such methods primarily address communication efficiency and privacy concerns by exchanging model updates instead of data. However, federated learning typically assumes that each participating device processes its entire local dataset during training, and thus does not directly address the computational burden associated with large or redundant datasets on edge devices. In contrast to these model- and system-centric methods, our work adopts a data-centric perspective, focusing on reducing the cost of learning by limiting the number of training samples processed on each edge device.\subsection{Dataset Pruning and Coreset Selection}

The primary objective of dataset pruning and coreset construction is to distill a large training corpus into a compact subset that rigorously approximates the optimization trajectory of the full dataset~\cite{Toneva2019Forgetting,Coleman2020SVP}. Within the deep learning landscape, these techniques are pivotal for accelerating convergence and mitigating memory bottlenecks. Prevailing methodologies predominantly quantify sample importance via gradient norms~\cite{paul2021EL2N}, loss magnitudes~\cite{INFOBATCH}, influence functions~\cite{zhang2024TTDS}, or second-order Hessian information~\cite{Coleman2020SVP,zhang2024TTDS}. While theoretically robust, the computationally prohibitive nature of these operations—often necessitating full dataset training, iterative backpropagation, or expensive matrix inversions—renders them ill-suited for resource-constrained edge environments where energy and thermal budgets are strictly capped.

Although recent heuristics utilizing early-training signals~\cite{paul2021EL2N} or curriculum-based metrics have demonstrated that significant data redundancy exists, these strategies are largely designed for centralized, server-grade infrastructure. They rarely account for the specific hardware limitations of mobile edge devices, such as volatile energy availability or heterogeneous compute profiles. Distinct from these server-centric paradigms, our framework leverages lightweight, first-order statistics derived from a truncated warm-up phase. By employing a deterministic selection mechanism, our approach enforces strict adherence to device-specific resource constraints while integrating seamlessly into standard on-device learning workflows.

\subsection{Data-Centric Efficiency Paradigms}

Complementary to pruning, the broader field of data-centric AI~\cite{data-centric} emphasizes optimizing the quality and curriculum of training data rather than solely refining model architectures. Prominent techniques include sample reweighting~\cite{shen2020sampleRw}, curriculum learning~\cite{bengio2009curriculum,huang2020curriclum}, and active learning~\cite{beluch2018activeL,li2024activeLsurvey}. Curriculum learning dynamically schedules samples by difficulty to stabilize training, whereas active learning iteratively queries labels for high-uncertainty data points. While effective in improving sample efficiency, these paradigms typically mandate continuous interaction, dynamic index updates, or complex scoring loops during the training process, introducing runtime overheads that are often prohibitive for lightweight edge nodes.

Conversely, our proposed method adopts a \emph{one-shot, deterministic} selection strategy executed purely locally prior to the main optimization phase. By fixing the training manifold early, we decouple data selection from gradient computation, thereby eliminating runtime scoring overhead and ensuring predictable resource consumption. This design choice is instrumental for practical edge deployments, where deterministic latency and energy efficiency are prerequisite operational requirements. Furthermore, this static reduction mechanism is strictly orthogonal to model-centric optimizations such as quantization or network pruning, allowing for synergistic deployment in extremely constrained environments. By addressing the efficiency bottleneck directly at the data level, our framework offers a complementary pathway to sustainable edge intelligence that operates independently of the specific model architecture employed. This orthogonality ensures that the proposed method can be universally applied as a foundational optimization layer across diverse edge learning ecosystems.


 \section{System Model}
\label{sec_system}

\begin{figure*}[t]
 \centering
 \includegraphics[width=7.0in]{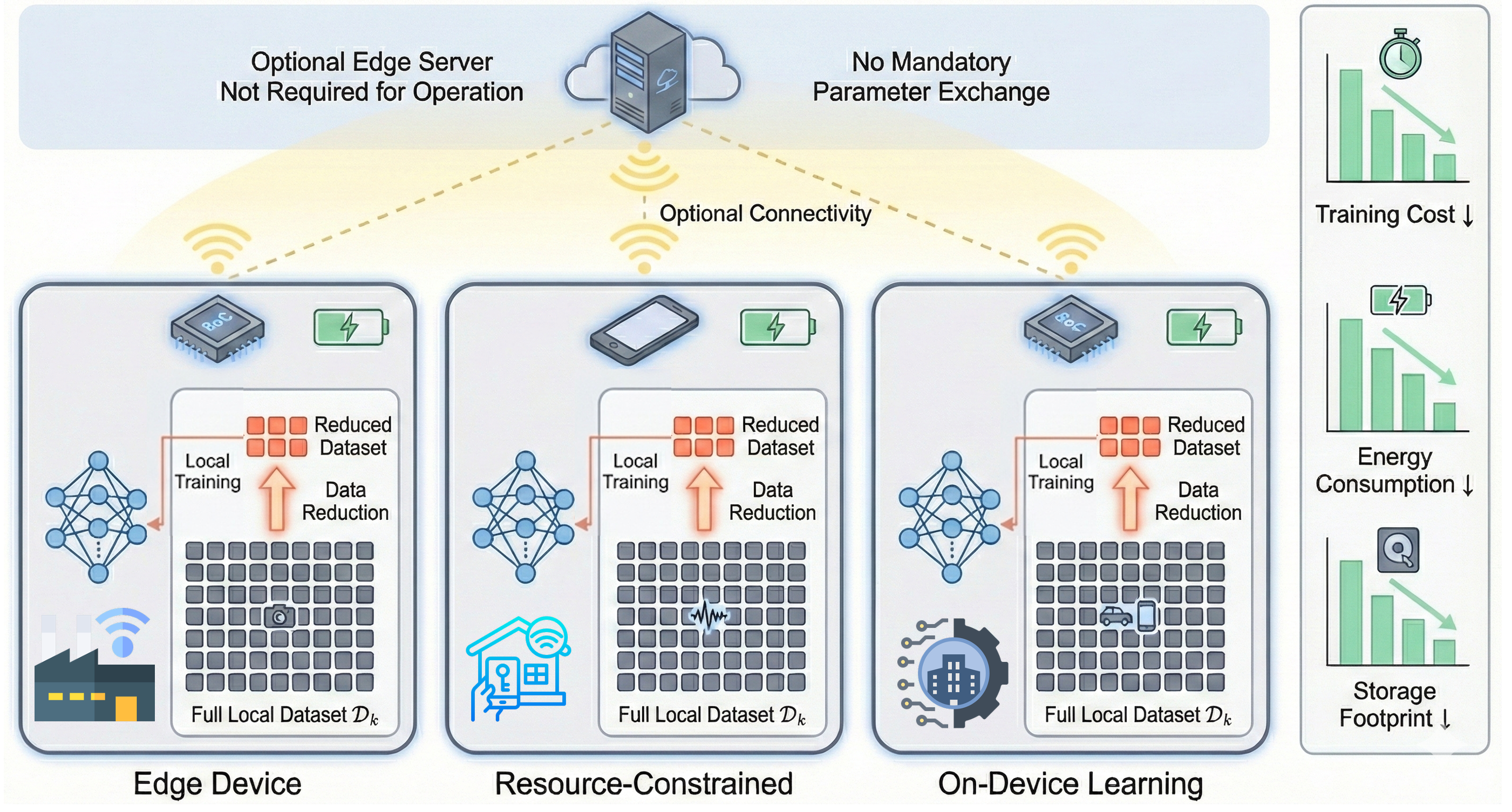}
 \caption{System model for lightweight edge learning via dataset pruning.}
 \label{fig:system_model}
\end{figure*}

We consider a decentralized edge learning framework in which multiple resource-constrained devices perform local model training or adaptation using their private data. This section details the learning setting, the proposed dataset pruning mechanism, the underlying learning dynamics, and the associated computation, energy, and storage cost models. The overall system architecture is illustrated in Fig.~\ref{fig:system_model}. Each edge device performs local dataset pruning and training independently. The edge server and communication channel are shown to illustrate potential extensibility to distributed or federated settings, but are not required by the proposed method.

\subsection{Edge Devices and Data Distribution}
Let $\mathcal{K} \triangleq \{1,2,\dots,K\}$ denote the set of participating edge devices. Each device $k\in\mathcal{K}$ possesses a local, labeled dataset
\begin{equation}
\mathcal{D}_k \triangleq \{(x_{k,n}, y_{k,n})\}_{n=1}^{N_k},
\end{equation}
where $x_{k,n} \in \mathcal{X}$ and $y_{k,n} \in \mathcal{Y}$ represent the input feature vector and corresponding label of the $n$-th sample, respectively, and $N_k = |\mathcal{D}_k|$ denotes the local dataset size. We assume that the samples on device $k$ are drawn from a device-specific distribution $\mathcal{P}_k \neq \mathcal{P}_j$ (for $k \neq j$), capturing the non-IID (Independent and Identically Distributed) nature of data in real-world edge environments.

\subsection{Learning Model and Objective}
We define a shared learning model $f(\cdot;\theta)$ parameterized by $\theta \in \mathbb{R}^d$, with a per-sample loss function $\ell(\cdot,\cdot)$. The objective for each device is to minimize its local empirical risk, defined as:
\begin{equation}
F_k(\theta) \triangleq \frac{1}{N_k} \sum_{n=1}^{N_k} \ell\!\left(f(x_{k,n};\theta), y_{k,n}\right).
\end{equation}
In this framework, devices execute training independently to adapt the model to their local environment, thereby avoiding the communication overhead associated with collaborative parameter exchange.

\subsection{Dataset Pruning Mechanism}
To mitigate the computational burden of training, each device $k$ constructs a pruned subset $\mathcal{S}_k \subseteq \mathcal{D}_k$ with cardinality $M_k \le N_k$. We introduce the \textit{pruning ratio} $\rho_k \triangleq M_k / N_k \in (0, 1]$, which quantifies the data reduction. Sample selection is formalized by binary indicator variables $s_{k,n} \in \{0,1\}$, subject to the cardinality constraint:
\begin{equation}
\sum_{n=1}^{N_k} s_{k,n} = M_k.
\end{equation}
Here, $s_{k,n}=1$ implies that the $n$-th sample is retained. Consequently, the device minimizes the \textit{pruned empirical risk}:
\begin{equation}
\tilde{F}_k(\theta) \triangleq \frac{1}{M_k} \sum_{n=1}^{N_k} s_{k,n} \, \ell\!\left(f(x_{k,n};\theta), y_{k,n}\right).
\end{equation}

\subsection{Learning Dynamics}
Each device optimizes $\tilde{F}_k(\theta)$ via mini-batch stochastic gradient descent (SGD). At iteration $t$, device $k$ samples a mini-batch $\mathcal{B}_{k,t} \subseteq \mathcal{S}_k$ of size $b_k$ and updates the parameters:
\begin{equation}
\theta_{k,t+1} = \theta_{k,t} - \eta_{k,t}\, g_{k,t},
\end{equation}
where $\eta_{k,t}$ is the learning rate, and the stochastic gradient $g_{k,t}$ is given by:
\begin{equation}
g_{k,t} \triangleq \nabla_{\theta} \left( \frac{1}{|\mathcal{B}_{k,t}|} \sum_{(x,y)\in\mathcal{B}_{k,t}} \ell\!\left(f(x;\theta_{k,t}), y\right) \right).
\end{equation}
Let $E_k$ be the number of local epochs. The total number of gradient update steps is:
\begin{equation}
T_k \approx E_k \left\lceil \frac{M_k}{b_k} \right\rceil = E_k \left\lceil \frac{\rho_k N_k}{b_k} \right\rceil.
\end{equation}
This relation explicitly demonstrates that for a fixed batch size, the computational complexity $T_k$ scales linearly with the pruning ratio $\rho_k$.

\subsection{Resource Cost Models}

\textbf{Computation and Energy:}
Let $c(\theta)$ represent the average computational workload (in FLOPs) for a single forward-backward pass. The per-iteration cost is $\mathrm{FLOPs}_k^{\mathrm{iter}} \approx c(\theta)\, b_k$. Given the device's computing throughput $f_k$ (FLOPs/s), the total training latency is:
\begin{equation}
\tau_k^{\mathrm{train}} \approx \frac{T_k \, \mathrm{FLOPs}_k^{\mathrm{iter}}}{f_k}.
\end{equation}
Assuming a constant operational power $P_k$, the total training energy consumption is $E_k^{\mathrm{train}} = P_k \tau_k^{\mathrm{train}}$. Both metrics are directly proportional to $\rho_k$, highlighting the efficiency gains of pruning.

\textbf{Storage:}
Let $\bar{\sigma}_k$ denote the average storage size of a data sample. The storage footprint for the pruned dataset is:
\begin{equation}
S_k^{\mathrm{pruned}} = M_k \bar{\sigma}_k = \rho_k S_k^{\mathrm{full}},
\end{equation}
where $S_k^{\mathrm{full}}$ is the baseline storage requirement for $\mathcal{D}_k$.


\section{Problem Formulation}
\label{sec_problem}

Dataset pruning aims to reduce the resource cost of edge learning while preserving model performance. Based on the system model in Section~\ref{sec_system}, our objective is to jointly learn the model parameters $\theta$ and select pruned datasets $\{\mathcal{S}_k\}_{k=1}^{K}$ for all participating edge devices.

\subsection{Objective of Edge Learning}
\label{sec_objective_standard}

We consider an edge learning system in which models are trained or adapted under strict resource constraints regarding computation, memory, and energy. Each edge device performs local training using its own data. For analytical convenience, we define an aggregate empirical risk over all devices as
\begin{equation}
\min_{\theta \in \mathbb{R}^d}
F(\theta)
\triangleq
\sum_{k=1}^{K} p_k
\frac{1}{|\mathcal{D}_k|}
\sum_{(x,y)\in\mathcal{D}_k}
\ell\!\left(f(x;\theta), y\right),
\label{eq:full_objective}
\end{equation}
where $p_k = \frac{|\mathcal{D}_k|}{\sum_{j=1}^{K}|\mathcal{D}_j|}$ reflects the
relative data volume at device $k$, and $\ell(\cdot,\cdot)$ is a task-specific
loss function.
This aggregate objective characterizes the overall learning goal across devices; in practice, each device optimizes its local empirical risk independently. While~\eqref{eq:full_objective} provides strong learning performance, the associated training cost scales linearly with the dataset size, which is often impractical for resource-constrained edge devices.


\subsection{Dataset Pruning for Efficient Edge Learning}
\label{sec_pruning_def}

To mitigate these costs, each device $k$ selects a pruned subset $\mathcal{S}_k \subseteq \mathcal{D}_k$ with a target cardinality $|\mathcal{S}_k| = M_k \ll N_k$. We define binary selection variables $s_{k,n}\in\{0,1\}$ such that:
\begin{equation}
\mathcal{S}_k = \{(x_{k,n},y_{k,n}) \in \mathcal{D}_k \mid s_{k,n}=1\}, \quad \text{s.t.} \sum_{n=1}^{N_k} s_{k,n} = M_k.
\end{equation}
Training is subsequently performed on these subsets.
For notational convenience, we define an aggregate pruned empirical risk as
\begin{equation}
\tilde{F}(\theta)
\triangleq
\sum_{k=1}^{K} p_k
\frac{1}{M_k}
\sum_{(x,y)\in\mathcal{S}_k}
\ell\!\left(f(x;\theta), y\right),
\label{eq:pruned_objective}
\end{equation}
which preserves the structure of the full empirical risk while operating on
significantly fewer training samples.
This formulation is used to characterize the overall effect of pruning across
devices; actual training is carried out locally on each device.

\subsection{Resource-Aware Optimization Objective}
\label{sec_resource_opt}

The joint dataset pruning and edge learning problem can be formulated as a resource-aware optimization that captures the trade-off between learning performance and resource consumption across edge devices. Let $C(\{\mathcal{S}_k\})$ be a cost function aggregating training time $\tau_k$, energy consumption $E_k$, and storage $S_k^{\mathrm{pruned}}$:
\begin{equation}
C(\{\mathcal{S}_k\}) \triangleq \lambda_{\tau}\sum_{k}\tau_k(\mathcal{S}_k) + \lambda_{E}\sum_{k}E_k(\mathcal{S}_k) + \lambda_{S}\sum_{k}S_k^{\mathrm{pruned}}.
\label{eq:resource_cost_detail}
\end{equation}
The global objective is to minimize the weighted sum of learning risk and resource overhead:
\begin{equation}
\begin{aligned}
\min_{\theta,\{s_{k,n}\}} \quad & \tilde{F}(\theta) + C(\{\mathcal{S}_k\}) \\
\text{s.t.} \quad & s_{k,n} \in \{0,1\}, \quad \forall k,n, \\
& \sum_{n=1}^{N_k} s_{k,n} = M_k, \quad \forall k,
\end{aligned}
\label{eq:joint_problem_refined}
\end{equation}
where $\lambda_{\tau}, \lambda_E, \lambda_S \ge 0$ balance performance and efficiency.

\subsection{Practical Relaxation via Importance-Based Pruning}

Directly solving~\eqref{eq:joint_problem_refined} is a combinatorial and intractable problem. In practice, pruning is executed by assigning each sample an importance score $\alpha_{k,n}$ derived from early-stage training dynamics. The pruned subset is efficiently obtained by:
\begin{equation}
\mathcal{S}_k = \operatorname{Top}\text{-}M_k \left( \{\alpha_{k,n}\}_{n=1}^{N_k} \right),
\label{eq:topm_selection_final}
\end{equation}
retaining only the most informative samples.

\section{Proposed Method}
\label{sec_method}

This section presents an importance-based dataset pruning method designed for efficient edge learning. The proposed approach approximates the resource-aware optimization problem formulated in Section~\ref{sec_problem} by selecting compact yet highly informative subsets of local training data prior to the primary training phase.

The central paradigm is to mitigate the computational and energy overhead of edge learning by eliminating redundant or less informative samples, rather than modifying model architectures or optimization algorithms. Specifically, each training sample is assigned a quantitative importance score derived from lightweight statistics obtained during early-stage training dynamics. Given a prescribed pruning ratio $\rho_k$, only samples with the highest importance scores are retained. The model is subsequently trained using the resulting pruned dataset $\mathcal{S}_k$ on each edge device. This design ensures that efficiency gains are achieved purely through data reduction, allowing for seamless integration with existing edge learning pipelines and federated frameworks.

\subsection{Importance Score Computation}
\label{sec_method_importance}

To approximate the resource-aware optimization objective defined in Section~\ref{sec_problem}, we quantify the contribution of each training sample to the learning process via a localized importance score. This score is computed during a truncated warm-up phase, avoiding the prohibitive costs of full convergence or complex second-order approximations, thereby suiting the constraints of edge hardware.

Let $\alpha_{k,n} \in \mathbb{R}_{+}$ denote the importance score of the $n$-th training sample $(x_{k,n}, y_{k,n})$ on device $k$. During an initial warm-up phase, each device performs $T_0$ stochastic gradient descent (SGD) iterations on its local dataset $\mathcal{D}_k$ using standard configurations. The importance score is defined as the average training loss of the sample over this warm-up period:
\begin{equation}
\alpha_{k,n} \triangleq \frac{1}{T_0} \sum_{t=1}^{T_0} \ell\!\left(f(x_{k,n};\theta_{k,t}), y_{k,n}\right),
\label{eq:importance_score}
\end{equation}
where $T_0 \ll T_k$ denotes the number of warm-up iterations and $\theta_{k,t}$ represents the model parameters at iteration $t$.

This scoring mechanism is motivated by the observation that, during the nascent stages of training, samples that are more difficult to fit typically exhibit higher loss values and generate gradients with larger magnitudes. Such samples are more influential in determining the optimization trajectory, whereas samples with consistently low loss values are often redundant or easily learned. From an optimization standpoint, the importance score in \eqref{eq:importance_score} serves as a first-order proxy for a sample's contribution to the empirical risk minimization process. 

Significantly, this computation requires only forward passes during the warm-up phase and bypasses the need for influence functions or Hessian approximations. Consequently, the computational overhead remains negligible relative to the total training cost, ensuring the method's feasibility for practical edge deployment.
\subsection{Pruned Dataset Construction}
\label{sec_method_pruning}

Given the importance scores $\{\alpha_{k,n}\}_{n=1}^{N_k}$ computed on device $k$, the objective of pruned dataset construction is to identify a compact subset that maximizes learning utility under a prescribed resource budget. To this end, sample selection is performed deterministically based on the relative ranking of importance scores, effectively approximating the combinatorial problem defined in Section~\ref{sec_problem}.

For a target pruning ratio $\rho_k \in (0,1]$, the cardinality of the retained subset is defined as:
\begin{equation}
M_k \triangleq \lfloor \rho_k N_k \rfloor,
\label{eq:mk_definition}
\end{equation}
where $N_k$ denotes the size of the original local dataset $\mathcal{D}_k$. The pruned dataset is constructed by retaining the $M_k$ samples associated with the largest importance scores. 

Formally, let $\pi_k: \{1, \dots, N_k\} \to \{1, \dots, N_k\}$ be a permutation that sorts the importance scores in non-increasing order, such that $\alpha_{k, \pi_k(1)} \ge \alpha_{k, \pi_k(2)} \ge \dots \ge \alpha_{k, \pi_k(N_k)}$. We represent sample selection via binary variables $s_{k,n}\in\{0,1\}$, where $s_{k,n}=1$ indicates that the $n$-th sample on device $k$ is retained. The selection rule is given by:
\begin{equation}
s_{k,n} = \mathbb{I} \left( \alpha_{k,n} \ge \alpha_{k, \pi_k(M_k)} \right),
\label{eq:indicator_selection}
\end{equation}
where $\mathbb{I}(\cdot)$ is the indicator function and $\alpha_{k, \pi_k(M_k)}$ denotes the $M_k$-th order statistic of the score set. Equivalently, this can be expressed as:
\begin{equation}
s_{k,n} = \begin{cases} 1, & \text{if } n \in \{ \pi_k(1), \dots, \pi_k(M_k) \}, \\ 0, & \text{otherwise}. \end{cases}
\label{eq:topm_selection}
\end{equation}

In the presence of ties (i.e., multiple samples with identical importance scores), a consistent tie-breaking rule, such as index-based prioritization, is applied to ensure that the cardinality constraint is strictly satisfied:
\begin{equation}
\sum_{n=1}^{N_k} s_{k,n} = M_k.
\label{eq:cardinality_constraint}
\end{equation}
This constraint directly enforces the resource-aware boundaries defined in Eq.~\eqref{eq:resource_cost_detail}. The resulting pruned dataset on device $k$ is defined as $\mathcal{S}_k \triangleq \{(x_{k,n}, y_{k,n}) \in \mathcal{D}_k \mid s_{k,n}=1\}$. All subsequent training procedures operate exclusively on $\mathcal{S}_k$ while model architecture and hyperparameters remain fixed.

\textbf{Computational overhead:}
The construction of $\mathcal{S}_k$ requires sorting or partial selection over $\{\alpha_{k,n}\}_{n=1}^{N_k}$. This can be implemented with an average complexity of $\mathcal{O}(N_k \log M_k)$ using partial sorting or $\mathcal{O}(N_k)$ using the quickselect algorithm. As this is a one-shot preprocessing step and $M_k \le N_k$, the overhead is negligible compared to the iterations required for full model convergence.
\subsection{Edge Learning with Pruned Data}
\label{sec_method_learning}

Following the deterministic construction of the pruned dataset $\mathcal{S}_k$, the model training phase is executed exclusively on these selected samples. A key advantage of this data-centric approach is that it requires no modifications to the model architecture, the loss function $\ell(\cdot)$, or the underlying optimization algorithm. Consequently, all efficiency gains are derived solely from the reduction in data volume, ensuring seamless compatibility with standard machine learning libraries and existing edge-to-cloud pipelines.

\textbf{Local Training Objective:}
For each device $k \in \mathcal{K}$, the local learning objective is to minimize the empirical risk over the pruned subset $\mathcal{S}_k$:
\begin{equation}
\min_{\theta} \; \tilde{F}_k(\theta) \triangleq \frac{1}{M_k} \sum_{(x,y)\in\mathcal{S}_k} \ell\!\left(f(x;\theta), y\right),
\label{eq:pruned_objective_local}
\end{equation}
where $M_k = |\mathcal{S}_k|$ is the cardinality of the pruned set. This objective serves as a high-fidelity approximation of the full empirical risk $F_k(\theta)$. By retaining the most influential samples—those with high importance scores $\alpha_{k,n}$—the optimization trajectory on $\mathcal{S}_k$ is designed to closely track the trajectory of the full dataset while requiring a fraction of the computational footprint.

\textbf{Local Optimization Dynamics:}
Each device optimizes \eqref{eq:pruned_objective_local} using mini-batch stochastic gradient descent (SGD). At iteration $t$, the parameter update is given by:
\begin{equation}
\theta_{k,t+1} = \theta_{k,t} - \eta_{k,t} \nabla_{\theta} \left( \frac{1}{b_k} \sum_{(x,y)\in\mathcal{B}_{k,t}} \ell\!\left(f(x;\theta_{k,t}), y\right) \right),
\label{eq:sgd_pruned}
\end{equation}
where $\mathcal{B}_{k,t} \subseteq \mathcal{S}_k$ is a mini-batch of size $b_k$ sampled from the pruned data. Since the pruning process is a one-shot preprocessing step, the standard convergence guarantees of SGD remain intact. However, the search space is now restricted to the most informative data manifold, which often leads to faster early-stage convergence relative to the number of samples processed.

\textbf{Quantitative Training Cost Reduction:}
Let $E_k$ denote the number of local training epochs. The total number of SGD updates on device $k$ is:
\begin{equation}
T_k^{\mathrm{pruned}} \approx E_k \frac{M_k}{b_k} = \rho_k \left( E_k \frac{N_k}{b_k} \right) = \rho_k T_k^{\mathrm{full}},
\label{eq:pruned_steps}
\end{equation}
where $\rho_k = M_k / N_k$ is the pruning ratio. This equation demonstrates that the computational workload scales linearly with $\rho_k$. As characterized in the resource model in Section \ref{sec_system}, this reduction translates directly into proportional savings in training latency $\tau_k^{\mathrm{train}}$ and energy consumption $E_k^{\mathrm{train}}$. This linear predictability is vital for edge systems where resource scheduling must be performed under strict energy budgets.


Overall, edge learning with pruned data provides a lightweight yet powerful mechanism to achieve resource efficiency. By focusing computation on the most informative samples, the system achieves a favorable trade-off between model accuracy and the physical costs of on-device learning.

\SetKwProg{Score}{Phase 1 - Importance Scoring:}{}{}
\SetKwProg{Select}{Phase 2 - Deterministic Selection:}{}{}
\SetKwProg{Train}{Phase 3 - Pruned Training:}{}{}

\begin{algorithm}[!t]
\DontPrintSemicolon
\SetAlgoLined
\caption{Importance-Based Dataset Pruning for Edge Learning (Device $k$)}
\label{alg:edge_pruning}

\KwIn{Local dataset $\mathcal{D}_k=\{(x_{k,n},y_{k,n})\}_{n=1}^{N_k}$;
pruning ratio $\rho_k \in (0,1]$; warm-up iterations $T_0$;
mini-batch size $b_k$; local epochs $E_k$;
learning rate schedule $\{\eta_{k,t}\}$;
loss $\ell(\cdot,\cdot)$; model $f(\cdot;\theta)$.}
\KwOut{Pruned subset $\mathcal{S}_k$; trained parameters $\theta_k$.}

$M_k \leftarrow \lfloor \rho_k N_k \rfloor$ \tcp*{Target subset size}
Initialize $\theta_{k,0}$\;

\BlankLine
\Score{}{
Initialize score vector $\bm{\alpha}_k \leftarrow \mathbf{0}_{N_k}$\;
\For{$t \leftarrow 1$ \KwTo $T_0$}{
    Sample mini-batch $\mathcal{B}_{k,t} \subseteq \mathcal{D}_k$ with $|\mathcal{B}_{k,t}|=b_k$\;
    \tcp{Accumulate per-sample losses on observed mini-batch}
    \ForEach{$(x_{k,n},y_{k,n}) \in \mathcal{B}_{k,t}$}{
        $\alpha_{k,n} \leftarrow \alpha_{k,n} + \ell\!\left(f(x_{k,n};\theta_{k,t-1}),y_{k,n}\right)$\;
    }
    \tcp{Standard SGD update}
    $g_{k,t} \leftarrow \nabla_{\theta}\!\left(\frac{1}{b_k}\sum_{(x,y)\in\mathcal{B}_{k,t}}
    \ell(f(x;\theta_{k,t-1}),y)\right)$\;
    $\theta_{k,t} \leftarrow \theta_{k,t-1} - \eta_{k,t} g_{k,t}$\;
}
Normalize $\bm{\alpha}_k$ by the number of times each sample is observed\;
}

\BlankLine
\Select{}{
$\mathcal{I}_k \leftarrow \operatorname{Top}\text{-}M_k(\{\alpha_{k,n}\}_{n=1}^{N_k})$\;
$\mathcal{S}_k \leftarrow \{(x_{k,n}, y_{k,n}) \in \mathcal{D}_k \mid n \in \mathcal{I}_k\}$\;
}

\BlankLine
\Train{}{
$\theta_k \leftarrow \theta_{k,T_0}$ \tcp*{Continue from warm-up}
\For{$t \leftarrow T_0{+}1$ \KwTo $T_0 + E_k\lceil M_k/b_k\rceil$}{
    Sample mini-batch $\mathcal{B}_{k,t} \subseteq \mathcal{S}_k$ with $|\mathcal{B}_{k,t}|=b_k$\;
    $\theta_k \leftarrow \theta_k - \eta_{k,t}\nabla_{\theta}\!\left(\frac{1}{b_k}\sum_{(x,y)\in\mathcal{B}_{k,t}}
    \ell(f(x;\theta_k),y)\right)$\;
}
}
\Return{$(\mathcal{S}_k,\theta_k)$}\;
\end{algorithm}

\begin{figure*}
 \centering
\includegraphics[width=7.0in]{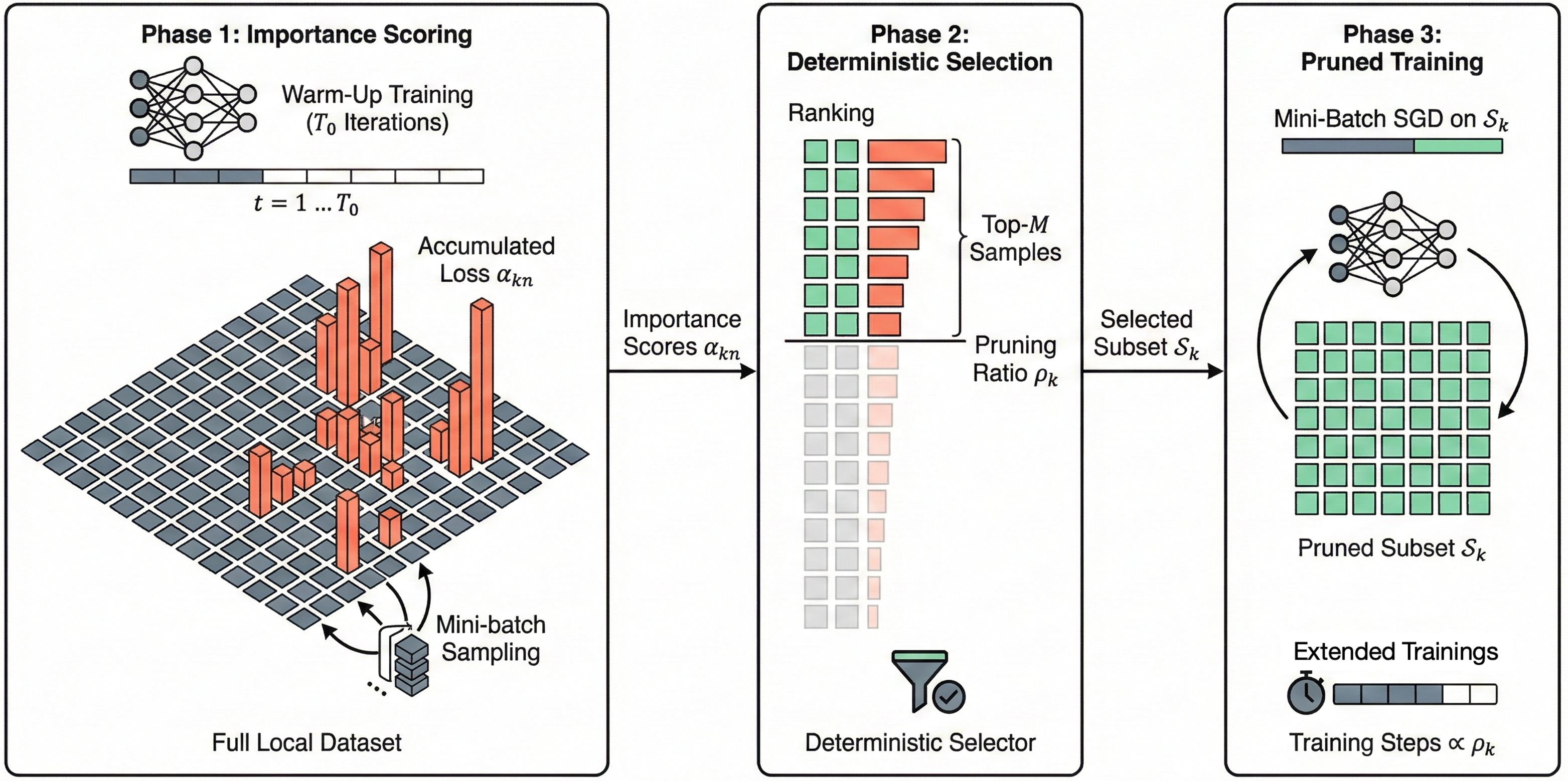}
 \caption{Three-Stage Importance-Based Dataset Pruning Pipeline}
 \label{fig:pipe}
\end{figure*}

Algorithm~\ref{alg:edge_pruning} summarizes the proposed importance-based dataset pruning pipeline executed independently on each edge device. The algorithm is structured into three distinct, modular stages (Fig.~\ref{fig:pipe}): (i) a truncated warm-up phase designed to compute per-sample importance scores using lightweight loss statistics; (ii) the deterministic construction of a pruned dataset $\mathcal{S}_k$ by selecting the top-ranked samples according to a prescribed pruning ratio $\rho_k$; and (iii) standard model training or adaptation performed exclusively on the resulting pruned subset.

Throughout this procedure, the model architecture, loss function, and optimization hyperparameters remain unchanged, ensuring that all efficiency gains are derived solely from strategic data reduction. By explicitly enforcing the cardinality constraint $M_k \le N_k$, the algorithm provides a practical approximation of the resource-aware optimization objective defined in Section~\ref{sec_problem}. Furthermore, because the scoring and selection steps are one-shot preprocessing operations, the pipeline incurs negligible computational overhead beyond the standard training process.

\subsection{Computational Complexity Analysis}
\label{sec_method_complexity}

We analyze the computational overhead of the proposed method on each edge device $k$, characterizing the complexity across three phases: (i) importance score computation, (ii) deterministic pruned dataset construction, and (iii) training on the resulting pruned subset. Complexity is evaluated in terms of local dataset size $N_k$, retained subset size $M_k = \rho_k N_k$, mini-batch size $b_k$, local epochs $E_k$, and the per-sample training workload $c(\theta)$ (FLOPs/sample) as defined in Section~\ref{sec_system}.

\textbf{A) Importance Score Computation:}
The importance scores are derived during a truncated warm-up phase of $T_0$ SGD iterations. Each iteration involves a forward and backward pass for a mini-batch of size $b_k$, resulting in a per-iteration cost of approximately $c(\theta)b_k$. The total computational workload for scoring is given by:
\begin{equation}
\mathrm{Cost}_{k}^{\mathrm{score}} = \sum_{t=1}^{T_0} c(\theta)b_k \approx T_0\,c(\theta)\,b_k.
\label{eq:cost_score}
\end{equation}
By setting $T_0$ to a small constant (e.g., a single epoch or a fixed number of iterations such that $T_0 \ll E_k N_k / b_k$), the scoring overhead becomes asymptotically negligible relative to the total training budget.

\textbf{B) Pruned Dataset Construction:}
The construction phase ranks samples based on $\{\alpha_{k,n}\}_{n=1}^{N_k}$ to select the top-$M_k$ elements. While full sorting incurs a complexity of $\mathcal{O}(N_k \log N_k)$, the use of a selection algorithm (e.g., Quickselect) reduces the expected complexity to linear time. The selection cost is denoted as:
\begin{equation}
\mathrm{Cost}_{k}^{\mathrm{select}} = \tilde{\mathcal{O}}(N_k),
\label{eq:cost_select}
\end{equation}
where $\tilde{\mathcal{O}}(\cdot)$ suppresses lower-order logarithmic factors. Because this step is a one-shot preprocessing operation executed once per training cycle, its cost is effectively amortized over the entire learning process.

\textbf{C) Training on the Pruned Dataset:}

On the pruned dataset $\mathcal{S}_k$, the total number of SGD update steps is $T_k^{\mathrm{pruned}} \approx \rho_k E_k \frac{N_k}{b_k}$. The total computational cost for training on the subset is:
\begin{equation}
\mathrm{Cost}_{k}^{\mathrm{train}} = T_k^{\mathrm{pruned}} c(\theta) b_k \approx \rho_k E_k N_k c(\theta).
\label{eq:cost_train}
\end{equation}
Compared to the full training cost $\mathrm{Cost}_{k}^{\mathrm{full}} \approx E_k N_k c(\theta)$, pruning achieves a multiplicative reduction:
\begin{equation}
\frac{\mathrm{Cost}_{k}^{\mathrm{train}}}{\mathrm{Cost}_{k}^{\mathrm{full}}} \approx \rho_k.
\label{eq:cost_reduction_ratio}
\end{equation}
This linear scaling provides a predictable mechanism for hardware resource management, allowing devices to adjust $\rho_k$ dynamically to meet strict energy and thermal envelopes.

\textbf{D) Overall Per-Device Complexity:}
The total computational footprint per device is the sum of its modular components:
\begin{equation}
\mathrm{Cost}_{k}^{\mathrm{total}} \approx \underbrace{T_0\,c(\theta)\,b_k}_{\text{Scoring}} + \underbrace{\tilde{\mathcal{O}}(N_k)}_{\text{Selection}} + \underbrace{\rho_k E_k N_k\,c(\theta)}_{\text{Pruned Training}}.
\label{eq:cost_total}
\end{equation}
In typical edge learning scenarios where $E_k N_k \gg T_0 b_k$, the workload is dominated by the pruned training term. Thus, the system achieves near-linear computational savings proportional to $1 - \rho_k$, with minimal preprocessing overhead. This analysis confirms the method's suitability for resource-constrained edge environments where maximizing performance-per-watt is critical.

\section{Performance Evaluation and Results}
\label{sec_results}

We evaluate the efficacy of the proposed dataset pruning framework as a purely local mechanism for reducing on-device training costs in edge learning systems. Our evaluation focuses on validating four key aspects: (1) \textbf{accuracy preservation}, ensuring importance-based pruning outperforms random selection and approaches full-data performance; (2) \textbf{linear cost scaling}, confirming that latency, FLOPs, and energy reduce proportionally with the pruning ratio $\rho$; (3) \textbf{minimal overhead}, verifying that the one-shot scoring phase is negligible compared to training savings; and (4) \textbf{robustness}, demonstrating effectiveness under both device heterogeneity and non-IID data distributions.

\subsection{Experimental Configuration}
\label{sec_setup}

To provide a comprehensive assessment, we conduct experiments on the CIFAR-10~\cite{krizhevsky2009cifar10} image classification benchmark using two distinct architectures: ResNet-34~\cite{he2016resnet}, representing high-capacity convolutional networks, and MobileNetV2~\cite{sandler2018mobilenetv2}, a lightweight architecture optimized for edge deployment. We simulate a network of $K=15$ edge devices, each performing fully local training on its private dataset. To mirror realistic edge environments, we evaluate both Independent and Identically Distributed (IID) settings and non-IID settings generated via a uniform distribution with concentration parameter $\beta$, which introduces significant class imbalance across devices.

All comparisons share an identical training protocol to ensure fairness, utilizing the Adam optimizer with a cosine annealing learning rate schedule and fixed batch sizes $b_k$. In our proposed method, importance scores $\alpha_{k,n}$ are computed during a truncated warm-up phase of $T_0$ iterations (as detailed in Section~\ref{sec_method_importance}), after which training proceeds exclusively on the retained subset $\mathcal{S}_k$. We benchmark our approach against two primary baselines: full-data training: The performance upper bound using the complete local dataset $\mathcal{D}_k$; and random pruning: A non-informative lower bound that retains $M_k$ samples selected uniformly at random.

All results are averaged over multiple independent trials to mitigate stochastic variance.

\begin{figure}[t]
 \centering
\includegraphics[width=3.5in]{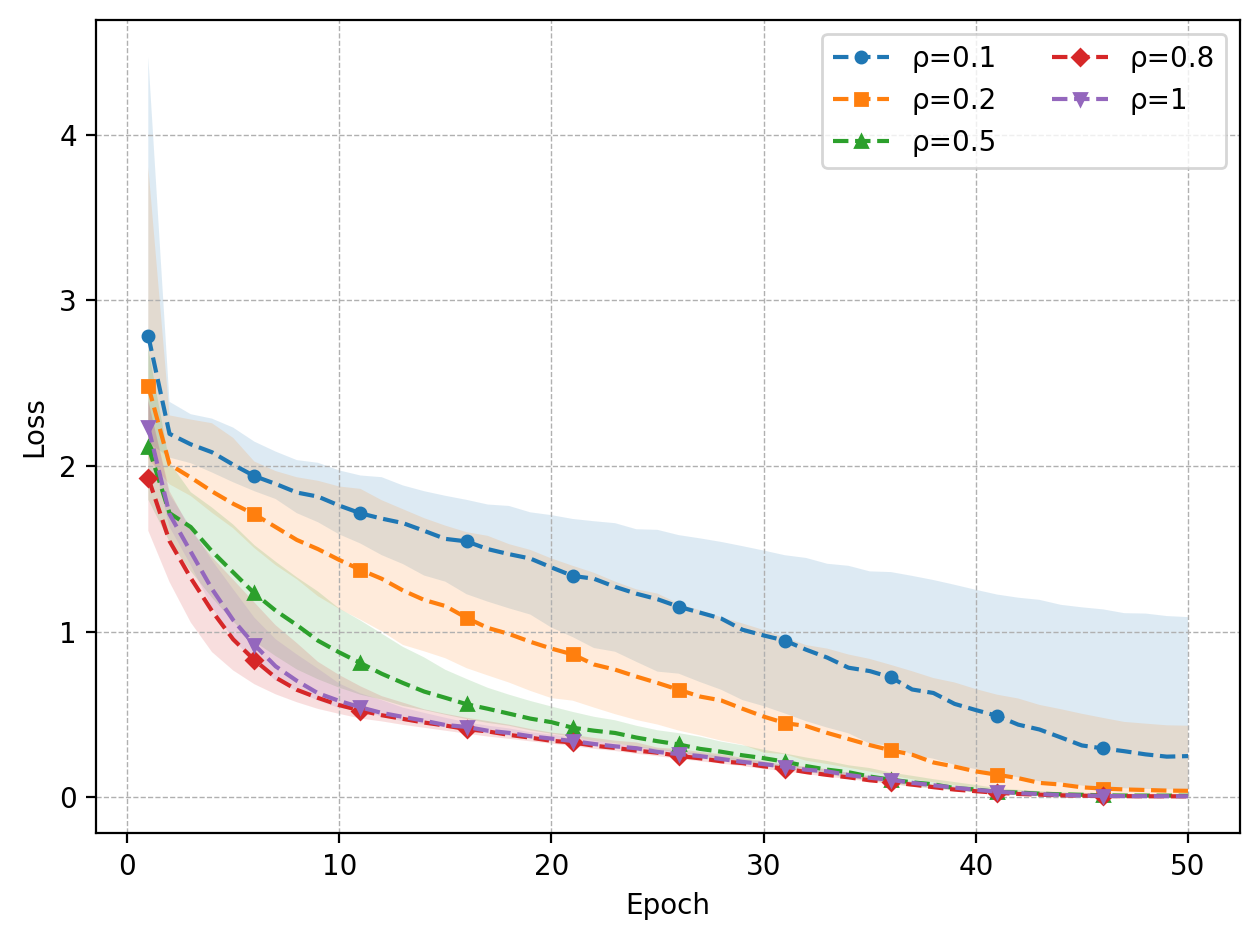}
 \caption{Training Loss}
 \label{fig:train_loss}
\end{figure}

\begin{figure}[t]
 \centering
\includegraphics[width=3.5in]{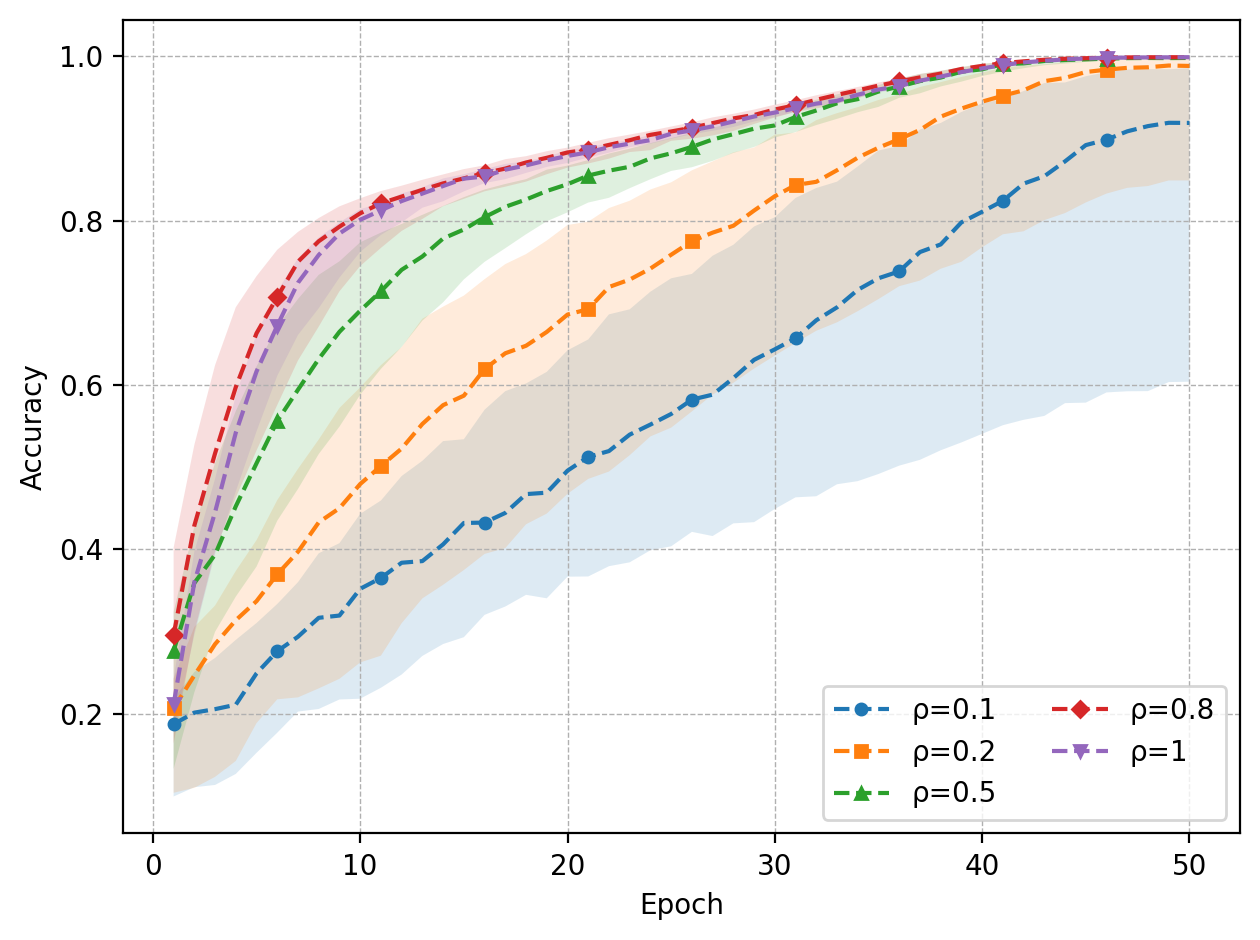}
 \caption{Training Accuracy}
 \label{fig:train_acc}
\end{figure}

\begin{figure}[t]
 \centering
\includegraphics[width=3.5in]{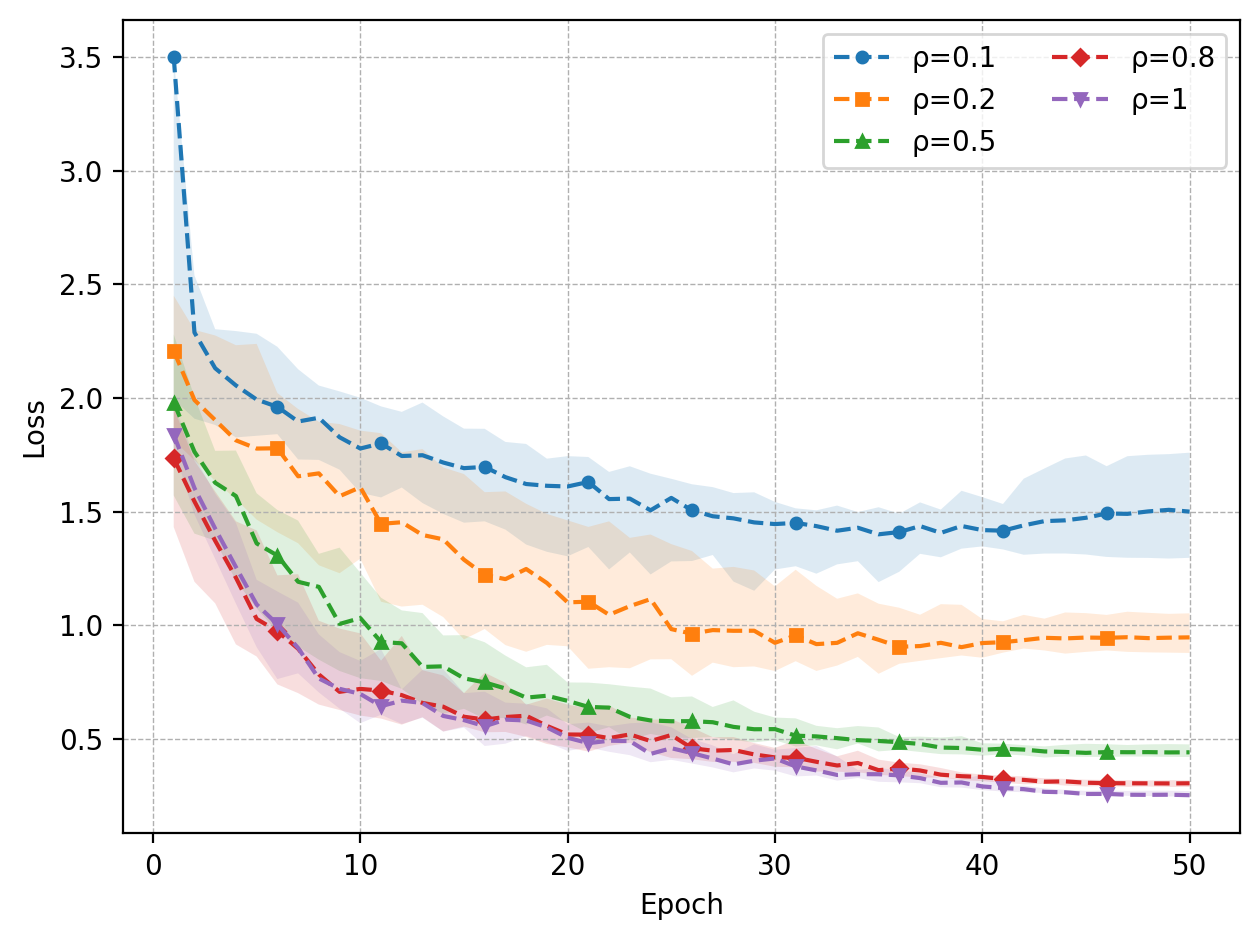}
 \caption{Testing Loss}
 \label{fig:test_loss}
\end{figure}

\begin{figure}[t]
 \centering
\includegraphics[width=3.5in]{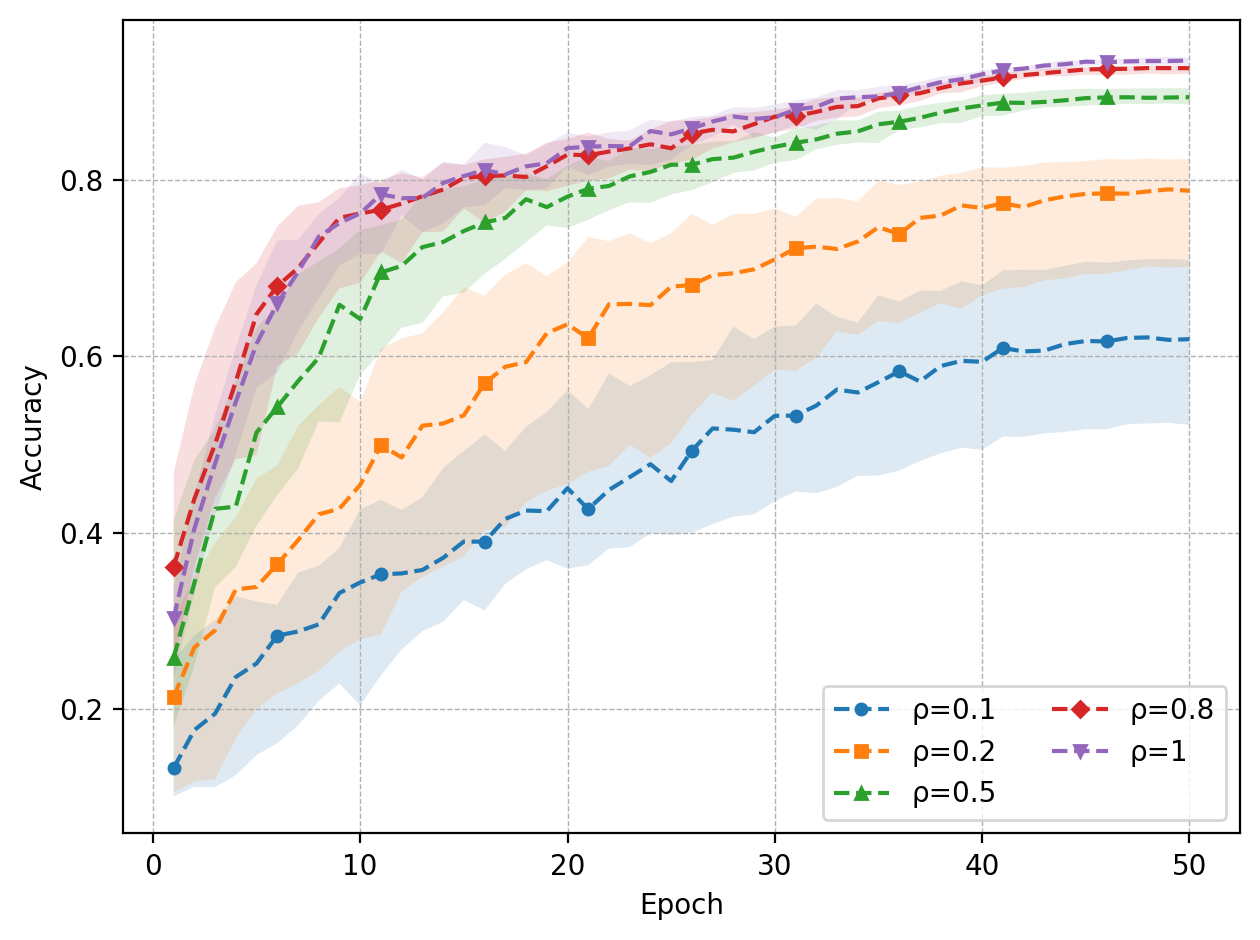}
 \caption{Testing Accuracy}
 \label{fig:test_acc}
\end{figure}

\subsection{Training Dynamics and Generalization Stability}
\label{sec_dynamics}
The training loss trajectories (Fig.~\ref{fig:train_loss}) reveal that importance-based pruning preserves the fundamental optimization landscape of the full-data baseline. While aggressive pruning (e.g., $\rho=0.1$) naturally incurs a higher initial loss due to the reduced cardinality of the training set, the convergence rate remains robust. Notably, for retention ratios $\rho \ge 0.5$, the loss curves exhibit smooth descent patterns that effectively mirror the full-data regime ($\rho=1.0$). This observation is corroborated by the training accuracy profiles in Fig.~\ref{fig:train_acc}, where the trajectory for $\rho=0.8$ is statistically indistinguishable from the baseline. This implies that approximately 20\% of the local dataset consists of redundant or low-information samples that contribute negligible marginal utility to the gradient descent process.

Regarding generalization, the testing loss evolution (Fig.~\ref{fig:test_loss}) highlights the stabilizing effect of the proposed importance scoring. While lower retention ratios introduce increased stochastic variance—manifested as wider confidence intervals in the early epochs—the method prevents catastrophic divergence. Crucially, as shown in Fig.~\ref{fig:test_acc}, the generalization gap remains minimal for moderate pruning ratios. The model trained with $\rho=0.5$ recovers the vast majority of the baseline accuracy, validating our hypothesis that importance-based selection successfully constructs a high-fidelity coreset. By filtering out redundant samples while retaining high-value examples, the pruning mechanism ensures that the edge device focuses its limited energy budget on refining the decision boundaries critical for robust generalization.

\subsection{Accuracy--Pruning Trade-offs}
\label{sec_acc_vs_rho}

\begin{figure}[t]
 \centering
\includegraphics[width=3.5in]{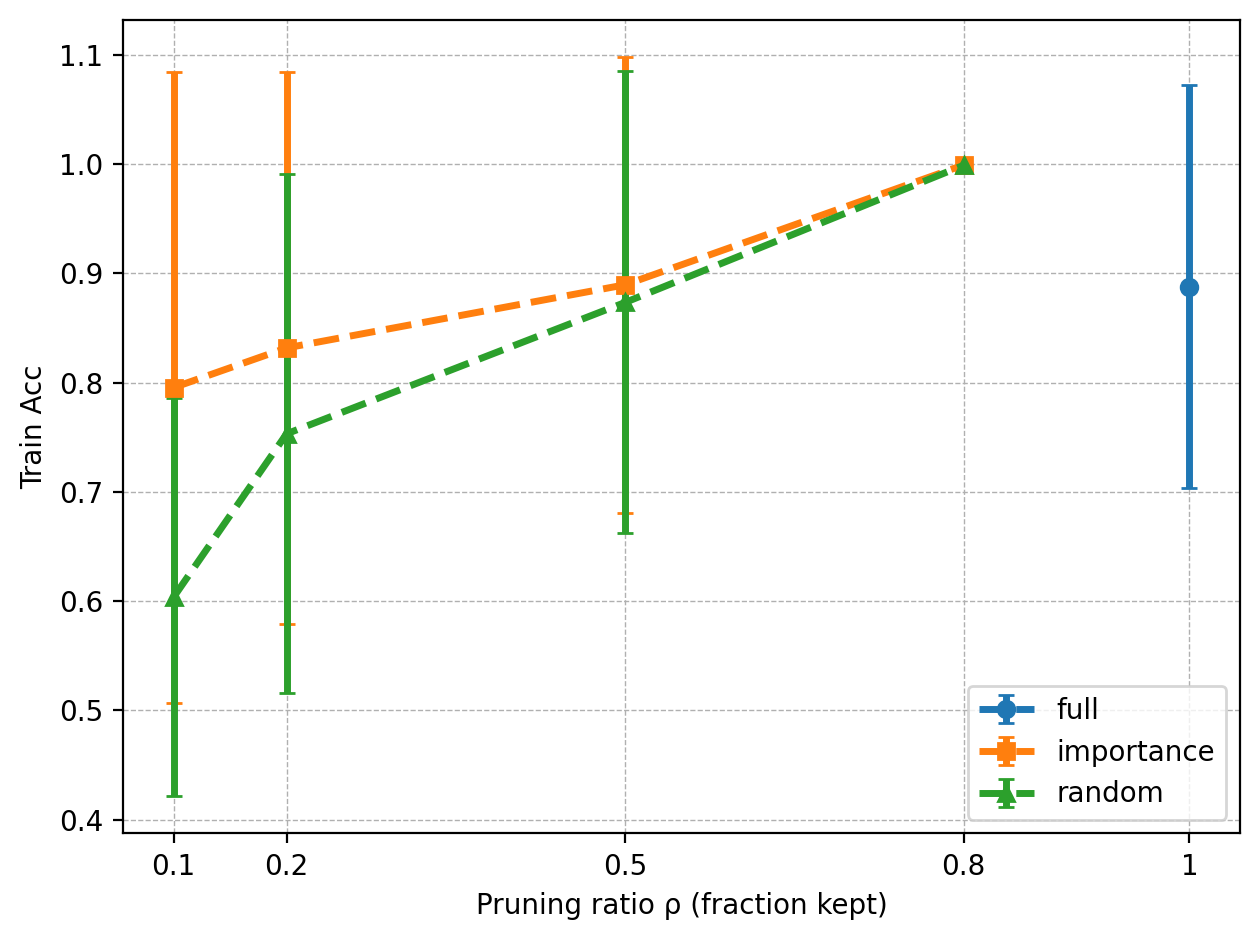}
 \caption{Training Accuracy vs Pruning Ratio}
 \label{fig:train_acc_vs_rho}
\end{figure}

\begin{figure}[t]
 \centering
\includegraphics[width=3.5in]{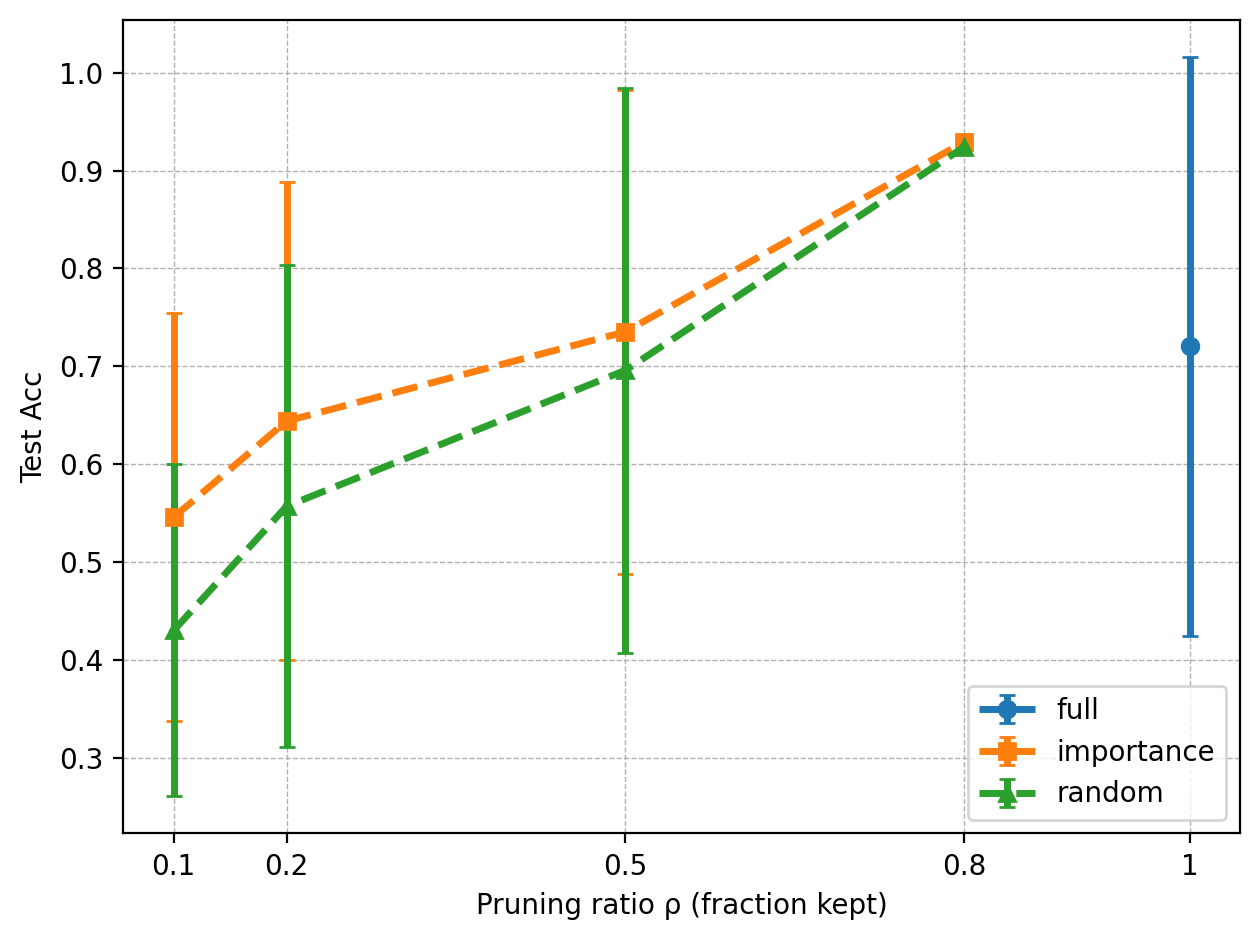}
 \caption{Testing Accuracy vs Pruning Ratio}
 \label{fig:test_acc_vs_rho}
\end{figure}
We next quantify the aggregate impact of dataset reduction on model performance by analyzing the final training and testing accuracy as a function of the pruning ratio \(\rho\), as summarized in Fig.~\ref{fig:train_acc_vs_rho} and Fig.~\ref{fig:test_acc_vs_rho}. As illustrated in Fig.~\ref{fig:train_acc_vs_rho}, the training accuracy exhibits a monotonic ascent with increasing \(\rho\). Crucially, the proposed importance-based pruning maintains a superior Pareto frontier compared to the random baseline. In the aggressive pruning regime (\(\rho \in \{0.1, 0.2\}\)), the performance gap is most pronounced; for instance, at \(\rho=0.2\), our method achieves a training accuracy gain of approximately \(10\%\) over uniform sampling. This indicates that the importance scoring mechanism successfully identifies and retains the high-utility samples required to drive optimization, whereas random selection discards critical gradient information.

Complementing these training dynamics, the testing accuracy trends in Fig.~\ref{fig:test_acc_vs_rho} further validate the efficacy of the proposed approach. Importance-based pruning consistently dominates the stochastic baseline across the entire spectrum of \(\rho\). The robustness of our method is particularly evident under severe resource constraints: at \(\rho=0.1\), it yields a test accuracy improvement of over \(12\%\) compared to random pruning. As the data budget increases to \(\rho=0.5\), the proposed method recovers nearly \(90\%\) of the peak accuracy, effectively matching the performance profile of the full-data baseline while utilizing only half the training samples. This convergence confirms that the pruned subsets serve as high-fidelity coresets that preserve the generalization capability of the original distribution.

\subsection{Accuracy Gap and Cost--Performance Trade-offs}
\label{sec_gap_tradeoff}

\begin{figure}[t]
 \centering
\includegraphics[width=3.5in]{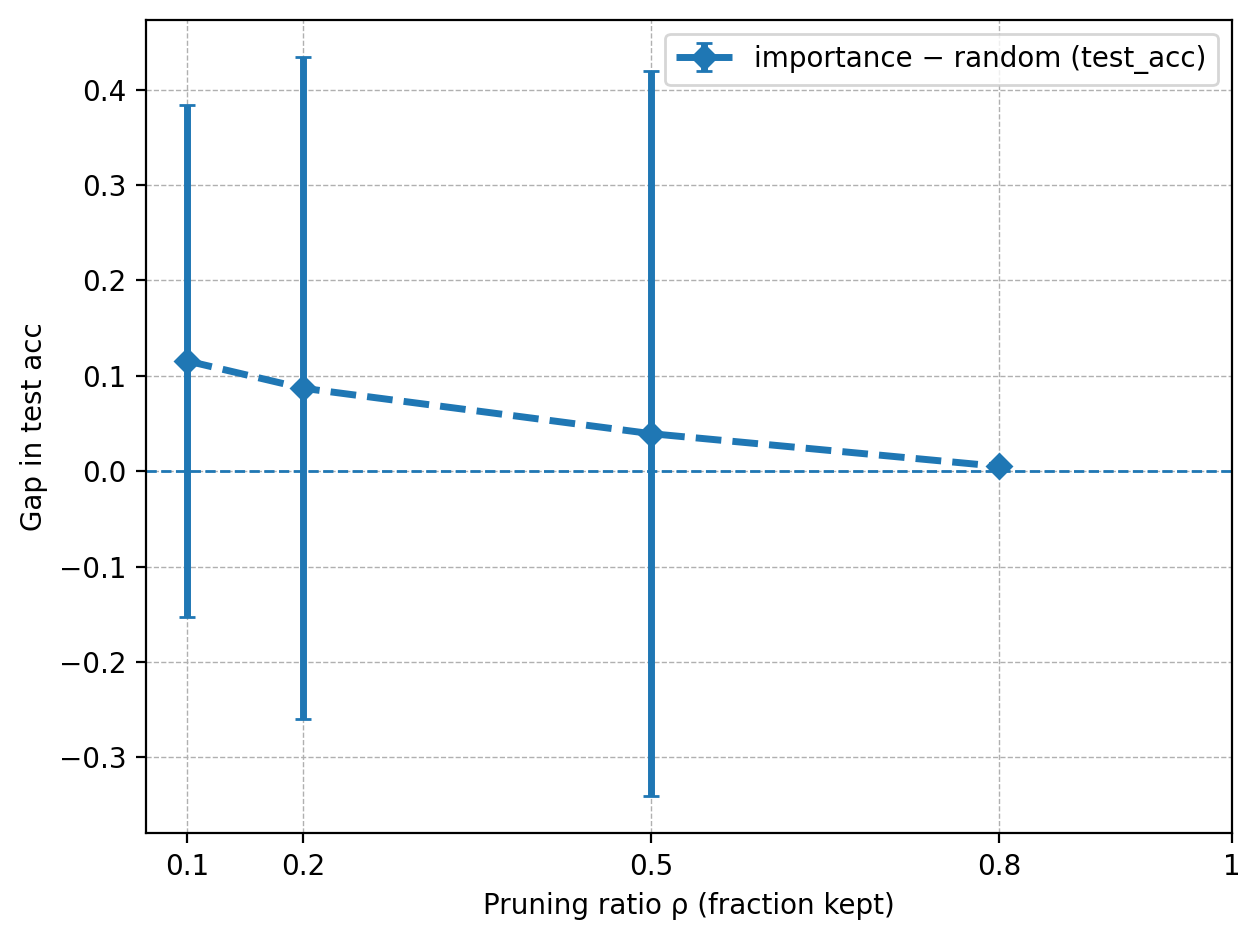}
 \caption{Accuracy Gap}
 \label{fig:acc_gap}
\end{figure}

\begin{figure}[t]
 \centering
\includegraphics[width=3.5in]{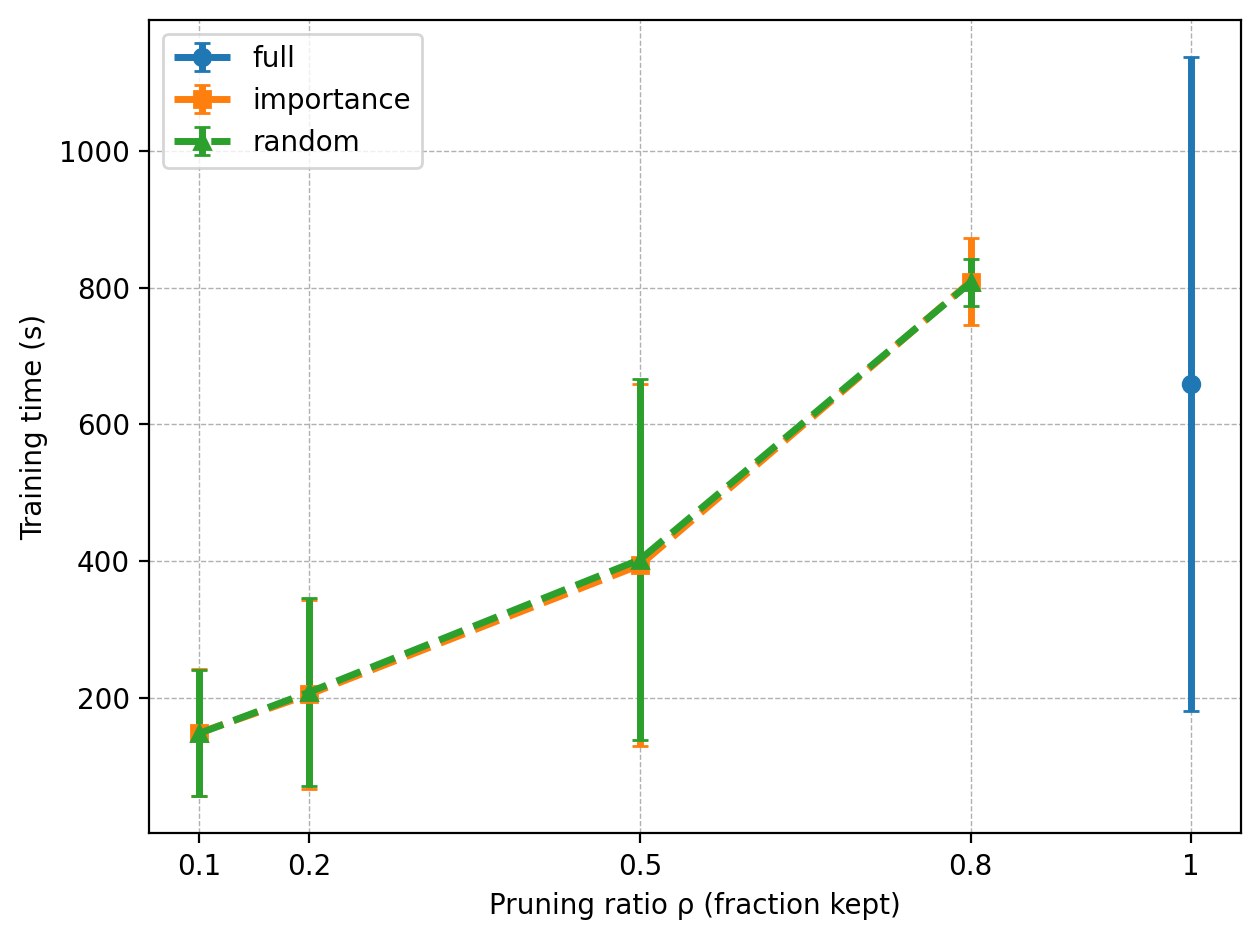}
 \caption{Training Time vs Pruning Ratio}
 \label{fig:train_time_vs_rho_rho}
\end{figure}

To further delineate the advantage of informed selection over stochastic sampling, Fig.~\ref{fig:acc_gap} quantifies the \emph{accuracy gap} between importance-based and random pruning as a function of the retained ratio $\rho$. This metric isolates the specific contribution of the importance scoring mechanism to the final model performance. Across all evaluated regimes, importance-based selection yields a strictly non-negative accuracy advantage. The benefits are most pronounced in the aggressive pruning regime ($\rho \leq 0.2$), where importance pruning outperforms random selection by approximately $10$--$15$ percentage points. This substantial margin confirms that under severe resource constraints, the proposed scoring effectively discriminates between critical data points and redundant samples, preventing the catastrophic performance drop observed in random baselines.

As the data budget increases ($\rho \to 0.8$), the accuracy gap asymptotically approaches zero. This behavior is expected, as the retained subset increasingly approximates the full distribution, naturally diminishing the variance between selection strategies. The convergence confirms that the proposed method introduces no inductive bias that might hinder performance when resources allow for milder pruning.

Complementing the accuracy analysis, Fig.~\ref{fig:train_time_vs_rho_rho} reports the wall-clock training latency. The results demonstrate that training time scales linearly with the retained data fraction for both methods, corroborating the theoretical complexity analysis in Section~\ref{sec_method_complexity}. Crucially, the latency profile of importance-based pruning is statistically indistinguishable from that of random pruning. This overlap indicates that the one-shot importance scoring and selection phase incurs negligible computational overhead compared to the iterative training process. Consequently, the proposed framework achieves a superior Pareto operating point: it delivers systematically higher accuracy—particularly in highly constrained environments—without incurring any penalty in runtime or energy consumption relative to naive baselines.

The experimental evaluation on CIFAR-10 demonstrates that the proposed importance-based pruning significantly outperforms random baselines, recovering near-optimal accuracy even when retaining only 50\% of the local training data. Furthermore, the results confirm a linear reduction in training latency and energy consumption proportional to the pruning ratio, validating that the method incurs negligible computational overhead. Finally, the approach exhibits strong robustness across diverse model architectures and non-IID data distributions, establishing its practical viability for resource-constrained edge environments.

\section{Conclusion}
\label{sec_conclusion}

In this paper, we established dataset pruning as a fundamental data-centric paradigm for enabling resource-efficient edge learning. By rigorously formulating the training process under strict computational, energy, and storage constraints, we demonstrated that the intelligent selection of a compact training subset allows edge devices to significantly reduce resource consumption without compromising model generalization. Unlike model-centric compression techniques, our approach reduces the underlying data volume, offering gains that are orthogonal to and combinable with lightweight architectures. We proposed a lightweight, importance-based pruning framework that leverages early-stage training dynamics to identify and retain high-utility samples. This mechanism enables the deterministic construction of local datasets that strictly adhere to device-specific resource budgets. Our evaluation confirms that this strategy yields a near-linear reduction in training latency and energy expenditure, directly proportional to the pruning ratio. Crucially, the method incurs negligible computational overhead and requires no modifications to the optimization algorithm, ensuring seamless deployability on standard edge hardware. Extensive experiments on image classification benchmarks validated that importance-based pruning consistently achieves a superior accuracy--efficiency trade-off compared to random selection, particularly in heterogeneous and non-IID environments. These findings position dataset pruning not merely as a heuristic, but as a robust, scalable building block for sustainable on-device intelligence. Future work will extend this framework to streaming data scenarios and investigate adaptive pruning mechanisms that dynamically adjust to time-varying energy harvesting profiles and real-time latency constraints.





\nocite{*}
\bibliographystyle{IEEEannot}
\bibliography{annot}

\begin{thebibliography}{10}
\providecommand{\url}[1]{#1}
\csname url@rmstyle\endcsname
\providecommand{\newblock}{\relax}
\providecommand{\bibinfo}[2]{#2}
\providecommand\BIBentrySTDinterwordspacing{\spaceskip=0pt\relax}
\providecommand\BIBentryALTinterwordstretchfactor{4}
\providecommand\BIBentryALTinterwordspacing{\spaceskip=\fontdimen2\font plus
\BIBentryALTinterwordstretchfactor\fontdimen3\font minus \fontdimen4\font\relax}
\providecommand\BIBforeignlanguage[2]{{%
\expandafter\ifx\csname l@#1\endcsname\relax
\typeout{** WARNING: IEEEtran.bst: No hyphenation pattern has been}%
\typeout{** loaded for the language `#1'. Using the pattern for}%
\typeout{** the default language instead.}%
\else
\language=\csname l@#1\endcsname
\fi
#2}}

\bibitem{Ale2024}
\BIBentryALTinterwordspacing
L.~Ale, N.~Zhang, S.~A. King, and D.~Chen, ``Empowering generative {AI} through mobile edge computing,'' \emph{Nature Reviews Electrical Engineering}, vol.~1, no.~7, pp. 478--486, 2024. [Online]. Available: \url{http://dx.doi.org/10.1038/s44287-024-00053-6}
\BIBentrySTDinterwordspacing


\bibitem{9955525}
D.~Gündüz, Z.~Qin, I.~E. Aguerri, H.~S. Dhillon, Z.~Yang, A.~Yener, K.~K. Wong, and C.-B. Chae, ``Beyond transmitting bits: Context, semantics, and task-oriented communications,'' \emph{IEEE Journal on Selected Areas in Communications}, vol.~41, no.~1, pp. 5--41, 2023.


\bibitem{9757749}
L.~Ale, S.~A. King, N.~Zhang, A.~R. Sattar, and J.~Skandaraniyam, ``{D3PG}: Dirichlet {DDPG} for task partitioning and offloading with constrained hybrid action space in mobile-edge computing,'' \emph{IEEE Internet of Things Journal}, vol.~9, no.~19, pp. 19\,260--19\,272, 2022.


\bibitem{sun2024EC}
H.~Sun, Y.~Zhou, H.~Zhang, L.~Ale, H.~Dai, and N.~Zhang, ``Joint optimization of caching, computing, and trajectory planning in aerial mobile edge computing networks: An maddpg approach,'' \emph{IEEE Internet of Things Journal}, vol.~11, no.~24, pp. 40\,996--41\,007, 2024.


\bibitem{Iterative_Pruning2017}
Z.~Liu, J.~Li, Z.~Shen, G.~Huang, S.~Yan, and C.~Zhang, ``Learning efficient convolutional networks through network slimming,'' in \emph{2017 IEEE International Conference on Computer Vision (ICCV)}, 2017, pp. 2755--2763.


\bibitem{10633894_pruning}
H.~Geng, Y.~Liu, Y.~Zheng, L.~L. Zhang, J.~Sun, Y.~Wang, Y.~Wang, G.~Sun, M.~Yang, T.~Cao, and Y.~Liu, ``Pruneaug: Bridging dnn pruning and inference latency on diverse sparse platforms using automatic layerwise block pruning,'' \emph{IEEE Transactions on Computers}, vol.~73, no.~11, pp. 2576--2589, 2024.


\bibitem{shim2024snp}
K.~Shim, J.~Yun, and S.~Choi, ``Snp: Structured neuron-level pruning to preserve attention scores,'' in \emph{European Conference on Computer Vision}.\hskip 1em plus 0.5em minus 0.4em\relax Springer, 2024, pp. 90--104.


\bibitem{zhou2017incremental}
A.~Zhou, A.~Yao, Y.~Guo, L.~Xu, and Y.~Chen, ``Incremental network quantization: Towards lossless cnns with low-precision weights,'' in \emph{International Conference on Learning Representations}, 2017.


\bibitem{jacob2018quantization}
B.~Jacob, S.~Kligys, B.~Chen, M.~Zhu, M.~Tang, A.~Howard, H.~Adam, and D.~Kalenichenko, ``Quantization and training of neural networks for efficient integer-arithmetic-only inference,'' in \emph{Proceedings of the IEEE conference on computer vision and pattern recognition}, 2018, pp. 2704--2713.


\bibitem{qu2025automatic}
X.~Qu, D.~Aponte, C.~Banbury, D.~P. Robinson, T.~Ding, K.~Koishida, I.~Zharkov, and T.~Chen, ``Automatic joint structured pruning and quantization for efficient neural network training and compression,'' in \emph{Proceedings of the Computer Vision and Pattern Recognition Conference}, 2025, pp. 15\,234--15\,244.


\bibitem{szegedy2015Googlenet}
C.~Szegedy, W.~Liu, Y.~Jia, P.~Sermanet, S.~Reed, D.~Anguelov, D.~Erhan, V.~Vanhoucke, and A.~Rabinovich, ``Going deeper with convolutions,'' in \emph{Proceedings of the IEEE conference on computer vision and pattern recognition}, 2015, pp. 1--9.


\bibitem{howard2017mobilenets}
A.~G. Howard, M.~Zhu, B.~Chen, D.~Kalenichenko, W.~Wang, T.~Weyand, M.~Andreetto, and H.~Adam, ``Mobilenets: Efficient convolutional neural networks for mobile vision applications,'' \emph{arXiv preprint arXiv:1704.04861}, 2017.


\bibitem{zhang2018Shufflenet}
X.~Zhang, X.~Zhou, M.~Lin, and J.~Sun, ``Shufflenet: An extremely efficient convolutional neural network for mobile devices,'' in \emph{2018 IEEE/CVF Conference on Computer Vision and Pattern Recognition}, 2018, pp. 6848--6856.


\bibitem{sandler2018mobilenetv2}
M.~Sandler, A.~Howard, M.~Zhu, A.~Zhmoginov, and L.-C. Chen, ``Mobilenetv2: Inverted residuals and linear bottlenecks,'' in \emph{Proceedings of the IEEE conference on computer vision and pattern recognition}, 2018, pp. 4510--4520.


\bibitem{Toneva2019Forgetting}
\BIBentryALTinterwordspacing
M.~Toneva, A.~Sordoni, R.~T. des Combes, A.~Trischler, Y.~Bengio, and G.~J. Gordon, ``An empirical study of example forgetting during deep neural network learning,'' in \emph{International Conference on Learning Representations}, 2019. [Online]. Available: \url{https://openreview.net/forum?id=BJlxm30cKm}
\BIBentrySTDinterwordspacing


\bibitem{2020ECsurvey}
X.~Wang, Y.~Han, V.~C.~M. Leung, D.~Niyato, X.~Yan, and X.~Chen, ``Convergence of edge computing and deep learning: A comprehensive survey,'' \emph{IEEE Communications Surveys \& Tutorials}, vol.~22, no.~2, pp. 869--904, 2020.


\bibitem{Ale2022EC}
L.~Ale, N.~Zhang, X.~Fang, X.~Chen, S.~Wu, and L.~Li, ``Delay-aware and energy-efficient computation offloading in mobile-edge computing using deep reinforcement learning,'' \emph{IEEE Transactions on Cognitive Communications and Networking}, vol.~7, no.~3, pp. 881--892, 2021.


\bibitem{2020LowRank}
Y.~Idelbayev and M.~A. Carreira-Perpiñán, ``Low-rank compression of neural nets: Learning the rank of each layer,'' in \emph{2020 IEEE/CVF Conference on Computer Vision and Pattern Recognition (CVPR)}, 2020, pp. 8046--8056.


\bibitem{park2019relational}
W.~Park, D.~Kim, Y.~Lu, and M.~Cho, ``Relational knowledge distillation,'' in \emph{2019 IEEE/CVF Conference on Computer Vision and Pattern Recognition (CVPR)}, 2019, pp. 3962--3971.


\bibitem{Beyer2022Kd}
L.~Beyer, X.~Zhai, A.~Royer, L.~Markeeva, R.~Anil, and A.~Kolesnikov, ``Knowledge distillation: A good teacher is patient and consistent,'' in \emph{2022 IEEE/CVF Conference on Computer Vision and Pattern Recognition (CVPR)}, 2022, pp. 10\,915--10\,924.


\bibitem{FederatedL&EC}
S.~Wang, T.~Tuor, T.~Salonidis, K.~K. Leung, C.~Makaya, T.~He, and K.~Chan, ``Adaptive federated learning in resource constrained edge computing systems,'' \emph{IEEE Journal on Selected Areas in Communications}, vol.~37, no.~6, pp. 1205--1221, 2019.


\bibitem{XIAO2021107338}
\BIBentryALTinterwordspacing
Z.~Xiao, X.~Xu, H.~Xing, F.~Song, X.~Wang, and B.~Zhao, ``A federated learning system with enhanced feature extraction for human activity recognition,'' \emph{Knowledge-Based Systems}, vol. 229, p. 107338, 2021. [Online]. Available: \url{https://www.sciencedirect.com/science/article/pii/S0950705121006006}
\BIBentrySTDinterwordspacing


\bibitem{lim2020FederatedLinEC}
W.~Y.~B. Lim, N.~C. Luong, D.~T. Hoang, Y.~Jiao, Y.-C. Liang, Q.~Yang, D.~Niyato, and C.~Miao, ``Federated learning in mobile edge networks: A comprehensive survey,'' \emph{IEEE Communications Surveys \& Tutorials}, vol.~22, no.~3, pp. 2031--2063, 2020.


\bibitem{Coleman2020SVP}
C.~Coleman, C.~Yeh, S.~Mussmann, B.~Mirzasoleiman, P.~Bailis, P.~Liang, J.~Leskovec, and M.~Zaharia, ``Selection via proxy: Efficient data selection for deep learning,'' in \emph{International Conference on Learning Representations (ICLR)}, 2020.


\bibitem{paul2021EL2N}
M.~Paul, S.~Ganguli, and G.~K. Dziugaite, ``Deep learning on a data diet: Finding important examples early in training,'' \emph{Advances in neural information processing systems}, vol.~34, pp. 20\,596--20\,607, 2021.


\bibitem{INFOBATCH}
Z.~Qin, K.~Wang, Z.~Zheng, J.~Gu, X.~Peng, Z.~Xu, Z.~Daquan, L.~Shang, B.~Sun, X.~Xie, and Y.~You, ``Infobatch: Lossless training speed up by unbiased dynamic data pruning,'' in \emph{International Conference on Representation Learning}, B.~Kim, Y.~Yue, S.~Chaudhuri, K.~Fragkiadaki, M.~Khan, and Y.~Sun, Eds., vol. 2024, 2024, pp. 39\,723--39\,744.


\bibitem{zhang2024TTDS}
X.~Zhang, J.~Du, Y.~Li, W.~Xie, and J.~T. Zhou, ``Spanning training progress: Temporal dual-depth scoring (tdds) for enhanced dataset pruning,'' in \emph{Proceedings of the IEEE/CVF Conference on Computer Vision and Pattern Recognition}, 2024, pp. 26\,223--26\,232.


\bibitem{data-centric}
S.~Salehi and A.~Schmeink, ``Data-centric green artificial intelligence: A survey,'' \emph{IEEE Transactions on Artificial Intelligence}, vol.~5, no.~5, pp. 1973--1989, 2024.


\bibitem{shen2020sampleRw}
Z.~Shen, P.~Cui, T.~Zhang, and K.~Kunag, ``Stable learning via sample reweighting,'' in \emph{Proceedings of the AAAI Conference on Artificial Intelligence}, vol.~34, no.~04, 2020, pp. 5692--5699.


\bibitem{bengio2009curriculum}
Y.~Bengio, J.~Louradour, R.~Collobert, and J.~Weston, ``Curriculum learning,'' in \emph{Proceedings of the 26th annual international conference on machine learning}, 2009, pp. 41--48.


\bibitem{huang2020curriclum}
Y.~Huang, Y.~Wang, Y.~Tai, X.~Liu, P.~Shen, S.~Li, J.~Li, and F.~Huang, ``Curricularface: Adaptive curriculum learning loss for deep face recognition,'' in \emph{2020 IEEE/CVF Conference on Computer Vision and Pattern Recognition (CVPR)}, 2020, pp. 5900--5909.


\bibitem{beluch2018activeL}
W.~H. Beluch, T.~Genewein, A.~N{\"u}rnberger, and J.~M. K{\"o}hler, ``The power of ensembles for active learning in image classification,'' in \emph{Proceedings of the IEEE conference on computer vision and pattern recognition}, 2018, pp. 9368--9377.


\bibitem{li2024activeLsurvey}
D.~Li, Z.~Wang, Y.~Chen, R.~Jiang, W.~Ding, and M.~Okumura, ``A survey on deep active learning: Recent advances and new frontiers,'' \emph{IEEE Transactions on Neural Networks and Learning Systems}, vol.~36, no.~4, pp. 5879--5899, 2024.


\bibitem{krizhevsky2009cifar10}
A.~Krizhevsky, G.~Hinton, \emph{et~al.}, ``Learning multiple layers of features from tiny images,'' 2009.


\bibitem{he2016resnet}
K.~He, X.~Zhang, S.~Ren, and J.~Sun, ``Deep residual learning for image recognition,'' in \emph{Proceedings of the IEEE conference on computer vision and pattern recognition}, 2016, pp. 770--778.


\end{thebibliography}
\end{document}